\documentclass[10pt,twocolumn,letterpaper]{article}

\usepackage{cvpr}
\usepackage{times}
\usepackage{epsfig}
\usepackage{graphicx}
\usepackage{amsmath}
\usepackage{amssymb}
\usepackage{algorithm}
\usepackage{algpseudocode}

\usepackage[pagebackref=true,breaklinks=true,colorlinks,bookmarks=false]{hyperref}

\cvprfinalcopy 

\def\cvprPaperID{****} 

\begin{document}

\title{Geometric VLAD for Large Scale Image Search}

\author{Zixuan Wang$^{\star}$, Wei Di$^{\dagger}$, Anurag Bhardwaj$^{\dagger}$, Vignesh Jagadeesh$^{\dagger}$, Robinson Piramuthu$^{\dagger}$ \\
$^{\star}$ Dept.of Electrical Engineering, Stanford University, CA 94305 \\
$^{\dagger}$ eBay Research Labs, eBay Inc., San Jose, CA 95125 \\
{\tt\small zxwang@stanford.edu, wedi@ebay.com, anbhardwaj@ebay.com, vjagadeesh@ebay.com, rpiramuthu@ebay.com}}


\maketitle

\begin{abstract}
We present a novel compact image descriptor for large scale image search. Our proposed descriptor - Geometric VLAD (gVLAD) is an extension of VLAD (Vector of Locally Aggregated Descriptors) that incorporates weak geometry information into the VLAD framework. 
The proposed geometry cues are derived as a membership function over keypoint angles which contain evident and informative information but yet often discarded. A principled technique for learning the membership function by clustering angles is also presented. Further, to address the overhead of iterative codebook training over real-time datasets, a novel codebook adaptation strategy is outlined. Finally, we demonstrate the efficacy of proposed gVLAD based retrieval framework where we achieve more than $15\%$ improvement in mAP over existing benchmarks.
\end{abstract}

\section{Introduction}
\label{sec:introduction}
\begin{figure}[t]
\centering
\includegraphics[width=1\linewidth]{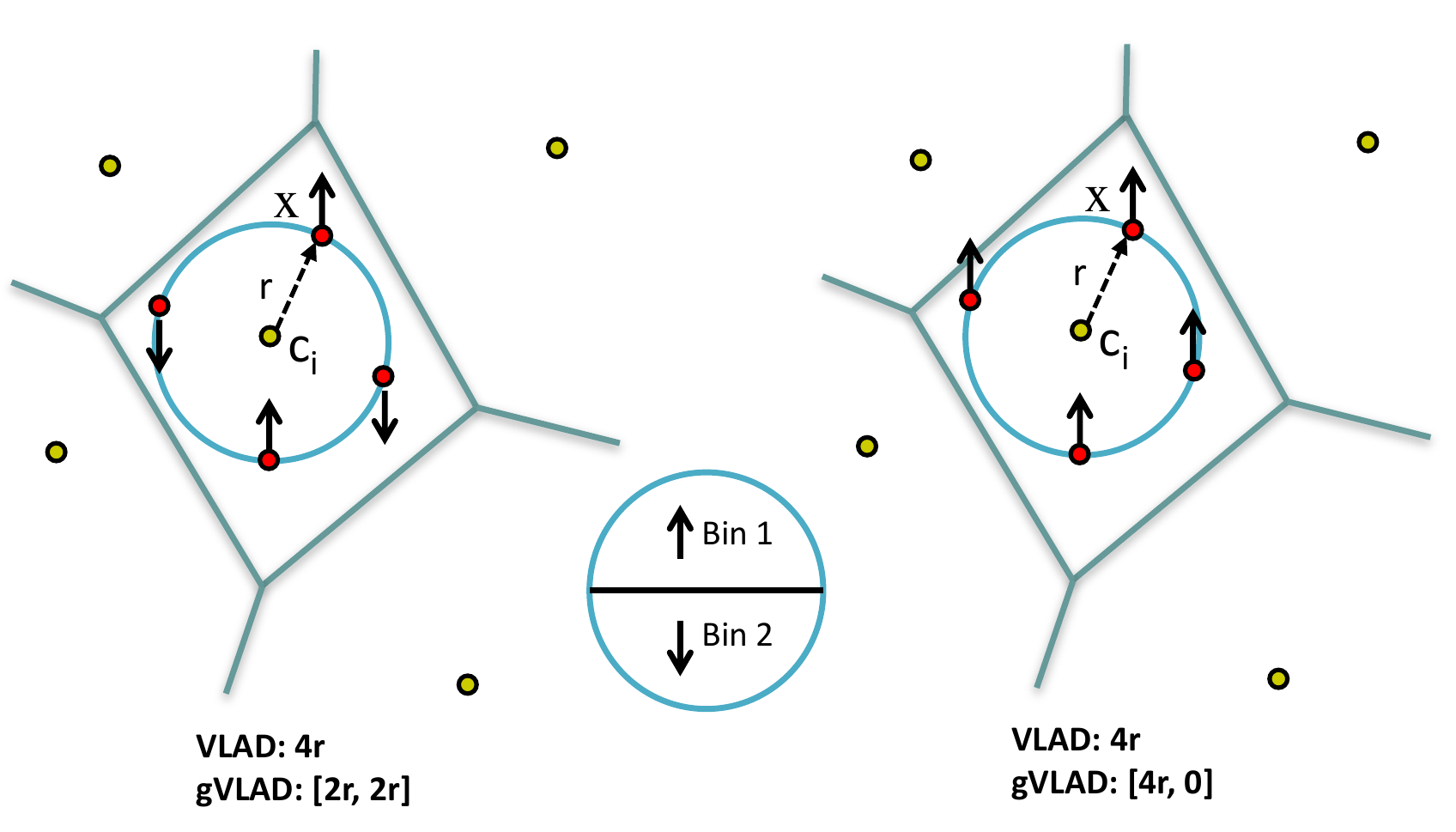} 
\caption{gVLAD motivation: A set of key points (denoted in red dot) locates in the feature space with same distance $r$ towards the centroid $c_i$, assuming they are of same feature descriptor. VLAD is unable to differentiate between the configurations shown in the left and right figures which differ only in the orientations of keypoints (depicted by arrow). However, by separating keypoints into two bins according to their dominant orientation, and measuring distance of points from each bin towards the centroid separately, the proposed gVLAD can successfully differentiate between the two configurations.}  
\label{fig:vlad_motivation}
\vspace{-2ex}
\end{figure}

Proliferation of large-scale image collections on web has made the task of efficient image retrieval challenging. 
Given a query image or region, the goal is to retrieve images of the same object or scene from a large scale database with high accuracy, efficiency and less memory usage. One of the core problems is how to concisely represent the visual information present in images. A number of methods have been proposed recently that address this issue from both computational efficiency as well as retrieval accuracy perspectives. However, there is a growing need for algorithms that can achieve reasonable trade-offs on both these aspects. Vector of Locally Aggregated Descriptors (VLAD)~\cite{jegou2010aggregating} proposed by J{\'e}gou et al. is one of the seminal contributions in this area as they show that compact and accurate VLAD representation is able to scale to billions of descriptors (by avoiding expensive hard disk operations) and still retain superior retrieval performance. However, one of the limitations of this representation is its inability to incorporate more keypoint level information that can potentially lead to enhanced performance. One such information is the dominant angle of the detected keypoint, also referred to as ``Keypoint Angle'', which is often discarded for the sake of obtaining rotational invariance in matches. 
A toy example is illustrated in Figure~\ref{fig:vlad_motivation}, in which VLAD is unable to differentiate between the configurations shown in the left and right figures where keypoints (red dots) differ only in their orientations, while having same descriptor representation and distance in the feature space towards the centroid $c_i$. 
Thus, we hypothesize that keypoint angles provide useful geometric cues which can be very useful for matching images. Integrating this information in a principled way can substantially improve the performance of existing VLAD based representation. In this paper, we present Geometric VLAD (gVLAD) that strengthens the VLAD representation by incorporating weak geometric cue in form of keypoint angles. 

Our contributions in this paper are as follows: 
\vspace{-1ex}
\begin{itemize} \itemsep0em
\item  \textbf{Angle Binning Based VLAD:} We propose a novel formulation of gVLAD that incorporates low level keypoint angles in form of a membership function into the VLAD representation. 
\item  \textbf{Circular Preserved Angle Membership Learning:} We propose a simple but effective principled technique to learn the membership function of keypoint angles based on trigonometric transform and clustering in a fashion that preserves their circular distribution.
\item \textbf{Codebook Adaptation:} To eliminate the need of iterative codebook training for large scale real-world image collections, a codebook adaptation scheme is presented.
\item \textbf{Z-Score Normalization:} Z-score based normalization technique is proposed that outperforms existing normalization methods for VLAD-based representations.
\item \textbf{Superior New Benchmark Results:} State-of-the-art image retrieval performance of proposed framework over a number of existing retrieval benchmarks are achieved.
\end{itemize}

The paper is organized as follows. In section~\ref{sec:relatedwork}, we outline related work in large-scale image search and strategies of integrating geometric information into image representations. In section~\ref{sec:model}, we describe the geometric VLAD representation in detail. In section~\ref{sec:experiment}, we demonstrate the performance gain on \textit{Oxford}, \textit{Holidays} and \textit{Paris} benchmarks, as well as on extended large scale datasets. We conclude the paper and discuss the future work in section \ref{sec:conclusion}.

\section{Related Work}
\label{sec:relatedwork}
The Bag-of-Words (BoW) representation is one of the most widely used method for image retrieval~\cite{sivic2003video,philbin2007object}. It quantizes each local descriptor SIFT~\cite{lowe1999object} or SURF~\cite{bay2006surf}, to its nearest cluster center and encodes each image as a histogram over cluster centers also known as ``Visual Words''. Good retrieval performance is achieved with a high dimensional sparse BOW vector, in which case inverted lists can be used to implement efficient search. However, the search time grows quadratically as the number of images increase~\cite{chum2010large}.

To overcome this issue, the Fisher kernel based approach proposed by Perronnin \textit{et al}.~\cite{perronnin2010large} transforms an set of variable-sized independent samples into a fixed size vector representation. The samples are distributed according to a parametric generative model, in this case a Gaussian Mixture Model (GMM) estimated on a training set. A simplified version of Fishers kernels, the VLAD is proposed by J{\'e}gou \textit{et al}.~\cite{jegou2010aggregating, jegou2012aggregating}. It encodes the difference from the cluster center in a more direct manner, rather than the frequency assigned to the cluster. It requires less computation than Fisher kernels but can achieve comparable retrieval performance.

However, most of existing methods ignore the geometric information present in images. 
Spatial re-ranking~\cite{philbin2007object} is usually used as a geometric filter to remove unrelated images from retrieval results. However, due to its expensive computation it is applied only to top ranked images for re-ranking. The spatial pyramid~\cite{lazebnik2006beyond} is a simple extension of the BOW representation which partitions the image into increasingly fine sub-regions and computes histograms of local features found inside each sub-region. It shows improved performance on scene classification tasks. The weak geometric consistency constraints (WGC)~\cite{jegou2008hamming} uses angle and scale information from key points to verify the consistency of matching descriptors. It can improve the retrieval performance significantly. Recently, Zhang \textit{et al}.~\cite{zhang2011image} propose a technique to encode more spatial information through the geometry-preserving visual phrases (GVP) which requires a pair of images to obtain geometric information. Chum \textit{et al}.~\cite{chum2009geometric} propose geometric min-hash, which extends min-hash by adding local spatial extent to increase the discriminability of the descriptor. It can be used for nearly duplicate image detection but has not achieved the state-of-the art performance in retrieval.

\begin{figure}[ht]
\centering
\begin{tabular}{cc}
\includegraphics[width=.47\linewidth]{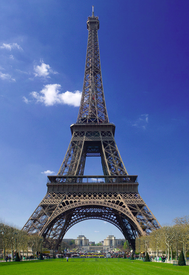}  &
\includegraphics[width=.47\linewidth]{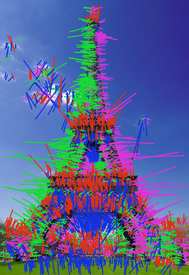} \\
\end{tabular}
\begin{tabular}{cc}
\includegraphics[width=.47\linewidth]{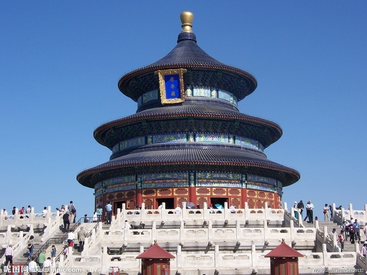}  &
\includegraphics[width=.47\linewidth]{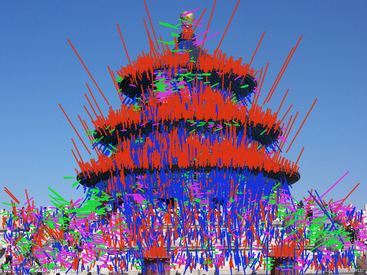} \\
\end{tabular}  
\caption{Each figure on the left shows an input image. Each figure on the right shows the detected key points in the image. Keypoints are grouped into four bins based on their angles represented by the direction of the line, and is colored by four unique colors. Length of the line corresponds to scale. Note that each image has a distinct representation based on the orientation of key points which suggests that this information can be potentially useful in image representation.}
\label{fig:objExperiments}
\end{figure}

\section{Proposed Framework}
\label{sec:model}
In this section, we introduce the Geometric VLAD (gVLAD) to improve retrieval performance by incorporating low level angle information from the key points into VLAD framework.

\subsection{Geometric VLAD}
\label{sec:gvlad}
Let us represent the local descriptor $\mathbf{x}$ to be $d$-dimensional vector (e.g. SURF or SIFT descriptors). Codebook or visual words are denoted as $\mu = \left[\mu_1, \mu_2, \dots, \mu_K\right]$, where $K$ represents the size of the vocabulary. Let $NN(\mathbf{x})$ represent the nearest-neighbor function that maps an input descriptor $\mathbf{x}$ to its nearest visual word $i$ where $1 \leq i \leq K$. In the original VLAD~\cite{jegou2010aggregating, jegou2012aggregating}, to represent a given image, a set of local descriptors are extracted first. Then, the contribution of each visual word $v_i$ is defined by accumulating distances of all the descriptors that belong to the $i^{th}$ visual word $\mu_i$ as: 
\begin{equation}
v_i = \sum_{\mathbf{x}:NN(\mathbf{x})=i}  \mathbf{x} - \mu_i
\end{equation}

Such representation is further L2-normalized, and concatenated to form a vector representation with size $d \times K$ to represent each image. However, the above formulation suffers from the drawback that it is unable to incorporate extra descriptor level information such as angle which can be of very useful in providing a weak geometrical cue. Thus, we present a gVLAD representation which encodes such angle information of the descriptor into the VLAD framework for efficient image matching. To define gVLAD, we redefine a local descriptor by $\mathbf{x}^{\theta}$, where $\mathbf{x}$ still represents the appearance feature vector of the descriptor and $\theta$ represents the angle of the descriptor, i.e. the dominant angle of the keypoint. For example, in SIFT descriptor, the angle corresponds to the dominant direction of gradient within a local window. To model the distribution of angles, we introduce clustering idea and define a membership function over the angles as: $\psi(\theta(\mathbf{x})):0 \leq \theta < 2\pi \to \{1,2,\dots,M\}$, where $M$ denotes the total number of angular bins.

The gVLAD $v_i^j$ for $i^{th}$ of the $K$ visual words (feature bin) and $j^{th}$ of the $M$ angular bins can now be represented as:
\begin{equation}
\label{eqn:gvlad}
 v_i^j = \left\{ 
  \begin{array}{l l}
    \sum_{ \mathbf{x}^{\theta} : NN(\mathbf{x}) = i}  \mathbf{x}^{\theta} - \mu_i & \quad \text{if $\psi(\theta(\mathbf{x}))=j$}\\
    \mathbf{0}^d & \quad \text{if $\psi(\theta( \mathbf{x})) \neq j$}
  \end{array} \right.
\end{equation}
where $d$ is the dimension of feature vector of local descriptor $\mathbf{x}$. The contribution of each visual word $V_i$ in the geometric VLAD can now be written as combining individual contributions from each angle bin:

\begin{equation}
V_i = [v_i^1, v_i^2, \cdots, v_i^{M-1}, v_i^M]
\label{eqn:V_i}
\end{equation}
where $V_i$ is a row vector with size of $d\times M$. Our geometric VLAD (gVLAD) representation $\mathcal{V}$ is defined by accumulating contributions of from all $K$ visual words, and has $D$ dimensions: $D=K\times d \times M$.

\begin{equation}
\mathcal{V} = [V_1, V_2, \cdots, V_{K-1}, V_{K}]
\end{equation}

\subsection{Learning Membership Function - $\psi(\theta)$}
\label{sec:membership}

One principal way to learn the membership function $\psi(\theta)$ is to apply clustering over the angle distribution and find the appropriate membership assignments for each angle value among $M$ learned clusters. Typically angles have a circular distribution of in the range of $[0, 2\pi )$, whereas existing clustering algorithms that based on L2 distance such as $k$-means assume a Cartesian co-ordinate space for input data, and can not be applied directly. To address this issue, we propose to represent each keypoint as $(r, \theta)$, where $r$ is the radial coordinate. Since we are only interested in angles of key point $\theta$, we fix $r$ as an arbitrary number $r>0$. We now perform a non-linear transform from this polar co-ordinate to 2D Cartesian co-ordinate space using the trigonometric functions:
\begin{align}
x &= r \times \cos{\theta} \\
y &= r \times \sin{\theta}
\end{align}
\noindent Thus, each angle is mapped to a point $\mathbf{z} (\theta)=(x,y)$ in this 2-d space.
To learn the membership of function $\psi(\theta)$, we perform k-means clustering in this space satisfying:
\begin{equation}
{\arg \min}_{  \lbrace \mathbf{\alpha}_1, \ldots , \mathbf{\alpha}_M  \rbrace  } \sum_{i=1}^M \sum_{\mathbf{z}_j \in \Xi_i} { \| \mathbf{z}_j - \mathbf{\alpha}_i   \| }^2
\end{equation}
\noindent where $\mathbf{\alpha}_i$ is the cluster centroid by averaging all points in cluster set $\Xi_i$.
The membership of each angle $\theta$ can be estimated through:
\begin{equation}
\psi(\theta) = {\arg \min}_{i \in  \lbrace1,2, \ldots, M \rbrace  }  { \| \mathbf{z}(\theta) - \mathbf{\alpha}_i   \| }^2
\end{equation}

\subsection{Codebook Adaptation}
\label{sec:adaptation}
Most real-world image databases grow continuously which leads to frequent codebook training processes that are often desirable. We propose a simple codebook adaptation process that can adapt existing codebooks with incremental dataset and alleviate the need of frequent large-scale codebook training. Secondly, this technique also allows codebook training from diverse datasets as codebook trained on one dataset (i.e. Paris building images) can be adapted to retrieve images from another dataset (i.e. Flickr holidays images). To define our codebook adaptation, let us represent a source dataset of images $S$ where an initial codebook $\mu = \left[\mu_1, \mu_2, \dots, \mu_K\right]$ is trained. Given a new domain specific dataset $T$, our goal is to adapt $\mu$ to another domain specific codebook $\hat{\mu}$ given as:
\begin{align}
\hat{\mu_i} &= \frac{1}{N}\sum_{t=1}^N \gamma_i(t),  \mathbf{x}^{\theta}(t) \in T  \\
\text{where}~\gamma_i(t) &= \left\{ 
  \begin{array}{l l}
   \mathbf{x}^{\theta}(t) & \quad \text{ if $NN( \mathbf{x}^{\theta}(t))=\mu_i$}\\
    \mathbf{0}^d & \quad \text{if $NN(\mathbf{x}^{\theta}(t)) \neq \mu_i$}
  \end{array} \right.
\end{align}
where $N$ is the total number of descriptors in dataset $T$ and $\mathbf{x}^{\theta}(t)$ represents $t^{th}$ descriptor. 
In our experiment, the initial codebook $\mu$ is trained using the \textit{Paris} dataset. For all the other experiments on different datasets, $\hat{\mu_i}$ is used in conjunction with Equation~\ref{eqn:gvlad} to compute the representation of the geometric VLAD.

\subsection{gVLAD Normalization}
\label{sec:z-normalization}
Normalization is important to effectively and correctly measure distance between vector representation. Here we propose three stages of normalization. 
First, we use the intra-normalization~\cite{arandjelovicall2013all}, where the sum of residuals of each visual word $v_i^j$ is L2 normalized independently, where $1 \leq i \leq K$ and $1 \leq j \leq M$. This step is followed by inter-Z-score normalization across different visual words. 
Given a vector $X$, its Z-score normalization is computed as: $\frac{X-\mu}{\sigma}$, where $\mu$ and $\sigma$ represent the mean and standard deviation of $X$. Let's denote the $t^{th}$ entry of $V_i$ as $v_{i,t}$, where $V_i$ is defined in Equation~\ref{eqn:V_i}. 
We apply the inter-Z-score normalization on each $ \left[  v_{1,t}, v_{2,t}, \ldots, v_{i,t}, \ldots, v_{K,t} \right]$,
where $1 \leq  t \leq  {M \times d}$ and $1 \leq i \leq K $. 
At last, the gVLAD vector $\mathcal{V}$ is L2 normalized $\mathcal{V} := \mathcal{V} / { \| \mathcal{V}\| }_2$.

\subsection{PCA Whitening}
\label{sec:pca-whitening}
Given a large collection of images, the size of representation needs to be carefully considered so as to be feasible for practical real time retrieval. For instance, using only $256$ visual words with $64$ dimensional SURF descriptors and $4$ angle bins generates a feature representation of size $D=64\times 256 \times 4=65,536$. To achieve memory-efficient representation of this vector, we use standard PCA with pre-whitening as described in~\cite{jegou2012negative}. The PCA whitening matrix can be expressed in the form of:
\begin{equation}
\textbf{P} = \textbf{D}^{-1/2}\textbf{E}^{T}
\end{equation}
where $\textbf{E} \textbf{D} \textbf{E}^{T}=E \lbrace  \bar{\mathbf{V}} \bar{\textbf{V}}^{T}  \rbrace$ is the eigenvector decomposition of the covariance matrix of the (zero mean) data $\bar{\textbf{V}}$, where each row $\bar{\mathcal{V}_l} = \mathcal{V}_l - \mathcal{V}_0$, and $\mathcal{V}_0$ is the mean vector computed from all gVALD representation vectors.  
$\textbf{D}=\mathrm{diag}[d_1, d_2, \dots, d_D]$ is the $D \times D $ diagonal matrix containing the eigenvalues and 
$\textbf{E}=[e_1, e_2, \dots, e_D]$ is an orthogonal $N \times D $ matrix having the eigenvectors as columns. 

The obtained whitened gVLAD representation is:
\begin{equation}
\tilde{\mathcal{V}_l} = \textbf{P}(:,1:\rho)^{T} \times \bar{\mathcal{V}_l}
\end{equation}
where $\rho$ is the number of eigenvectors to keep, i.e. the dimension of reduced feature vectors.  $\tilde{V}_l$ is then L2 normalized. The complete algorithm is outlined in Algorithm~\ref{alg:flow}.

\begin{algorithm}[htb]
	\caption{Computation of gVLAD descriptor}
	\begin{algorithmic}[1]
		\State{\textbf{S1:} Keypoint detection and descriptor: compute image descriptors $\mathbf{x}^\theta$, where $\mathbf{x}$ is the appearance vector, and $\theta$ represents the angle.}
		\State{\textbf{S2:} Generate visual vocabulary $\left[\mu_1, \mu_2, \dots, \mu_K\right]$ using $k$-means on all descriptors from training data.}
		\State{\textbf{S3:} Learning membership function $\psi(\theta)$ for each $\mathbf{x}^\theta$ \\\hspace{2cm} 
			   $ \arg \min_{i \in  \lbrace1,2, \ldots, M \rbrace  }  { \| \mathbf{z}(\theta) - \mathbf{\alpha}_i   \| }^2$ }
		\State{\textbf{S4:} Compute geometric VLAD $v_i^j$: \\
				$ v_i^j = \left\{ 
  \begin{array}{l l}
    \sum_{ \mathbf{x^{\theta}} : NN(\mathbf{x}) = i}  \mathbf{x}^{\theta} - \mu_i & \quad \text{if $\psi(\theta(\mathbf{x}))=j$}\\
    \mathbf{0}^d & \quad \text{if $\psi(\theta( \mathbf{x})) \neq j$}
  \end{array} \right.	$  \\
   \hspace{10mm} $ V_i = [v_i^1, v_i^2, \cdots, v_i^{M-1}, v_i^M] $  \\
   \hspace{12mm}$\mathcal{V} = [V_1, V_2, \cdots, V_{K-1}, V_{K}]$
  } 
		\State{\textbf{S5:} {Codebook Adaptation: \\\hspace{2cm} 
		$\hat{\mu_i} = \frac{1}{N}\sum_{t=1}^N \gamma_i(t), \mathbf{x}^{\theta}(t) \in T$}}
    \State{\textbf{S6:} Intra-normalization, Inter-Z-score normalization, and L2-normalization.}
		\State{\textbf{S7:} PCA whitening: \\\hspace{2cm}
		$\tilde{\mathcal{V}_l} = \textbf{P}(:,1:\rho)^{T} \times \bar{\mathcal{V}_l}$}
   \end{algorithmic}
  \label{alg:flow}
\end{algorithm}

\begin{figure*}[!ht]
  \begin{center}
    \includegraphics[width=.9\textwidth]{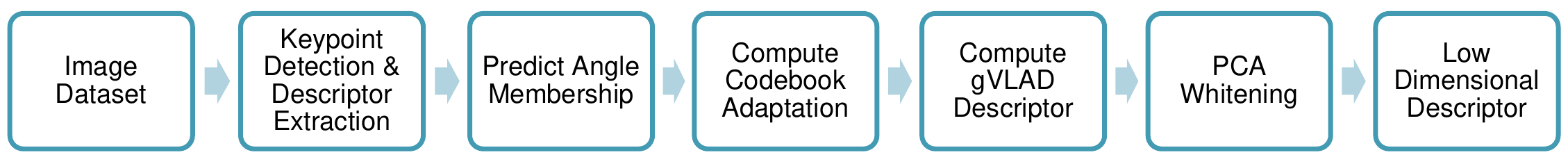}
  \end{center}
  \vspace{-0.3cm}
  \caption{The pipeline of the geometric VLAD (gVLAD) descriptor computation.}
  \label{fig:pipeline}
\end{figure*}

\section{Experiments \& Evaluations}
\label{sec:experiment}

\subsection{Benchmark Dataset}
\label{sec:dataset}
We evaluate the proposed approach on several public available benchmark datasets:
\textit{Oxford} buildings, \textit{Paris} and \textit{Holidays}. Large scale experiments are conducted on these datasets by adding $1$M Flickr images as distractors~\cite{jegou2008hamming}. For each of these datasets, performance is measured by mean average precision (mAP) over a set of pre-defined queries and their annotated ground truth matches.

\noindent\textbf{Holidays Dataset:}
\textit{Holidays} dataset~\cite{jegou2008hamming} contains $1491$ high resolution personal holiday photos with $500$ annotated queries. For large scale experiments, $1$ million Flickr images are added to it to create \textit{Holidays + Flickr 1M} dataset. About 5\%-10\% of the images in holiday dataset have orientations which are unnatural for human observer~\cite{perd2009efficient}. We manually rotate these images to create \textit{Rotated Holidays} dataset.

\noindent\textbf{Oxford Dataset:}
This dataset, \textit{Oxford 5K} contains $5062$ images of Oxford buildings gathered from Flickr~\cite{philbin2007object}. There are 55 query images each with a rectangular bounding box specifying the region of interest. To test large scale retrieval, it is firstly extended with a 100K Flickr images\footnote{\href{http://www.robots.ox.ac.uk/~vgg/data/oxbuildings/flickr100k.html}{http://www.robots.ox.ac.uk/~vgg/data/oxbuildings/flickr100k.html}} to create \textit{Oxford 105K } dataset. We further extend the dataset with 1 million Flickr image\footnote{\href{http://press.liacs.nl/mirflickr/}{http://press.liacs.nl/mirflickr/}} creating \textit{Oxford 5K + Flickr 1M} dataset.

\noindent\textbf{Paris Dataset:}
The Paris Dataset~\cite{philbin2008lost} \textit{Paris 6K} consists of $6412$ images collected from Flickr by searching for particular Paris landmarks. There are $60$ query images, each with a rectangular bounding box specifying the region of interest. We found that both 100K Flickr images and Flickr 1M images contains a large number of Paris landmarks, hence we do not extend the Paris dataset with Flickr images.

\subsection{Implementation Details}
\label{sec:implementation}

\noindent\textbf{Descriptor computation:}
The pipeline of computing gVLAD descriptor is characterized in Figure~\ref{fig:pipeline}. First, all images are resized to $1024\times 768$. We find that when using the original resolution of \textit{Holidays} images, the performance is inferior to the down-sampled images. We can also benefit from the speed when using smaller images. In \textit{Oxford} and \textit{Paris} datasets, bounding boxes are provided for queries. We only extract descriptors inside bounding boxes. We use the SIFT and SURF implementations in OpenCV\footnote{\href{http://opencv.org/}{http://opencv.org/}} to detect keypoints and extract descriptors. Each SIFT descriptor has 128 dimensions and each SURF descriptor has 64 dimensions. We find that VLAD based features have better performance using SURF keypoints and descriptors~\cite{bay2006surf} than SIFT keypoints and descriptors~\cite{lowe1999object}. In general, we observed about 10\% improvement using SURF as compared to SIFT. More details about the performance difference can be seen by comparing results in Table~\ref{tbl:root_sift_vlad} and Table~\ref{tbl:step-performance}. 

\begin{figure}[h]
\centering
\begin{tabular}{cc}
\includegraphics[width=.47\linewidth]{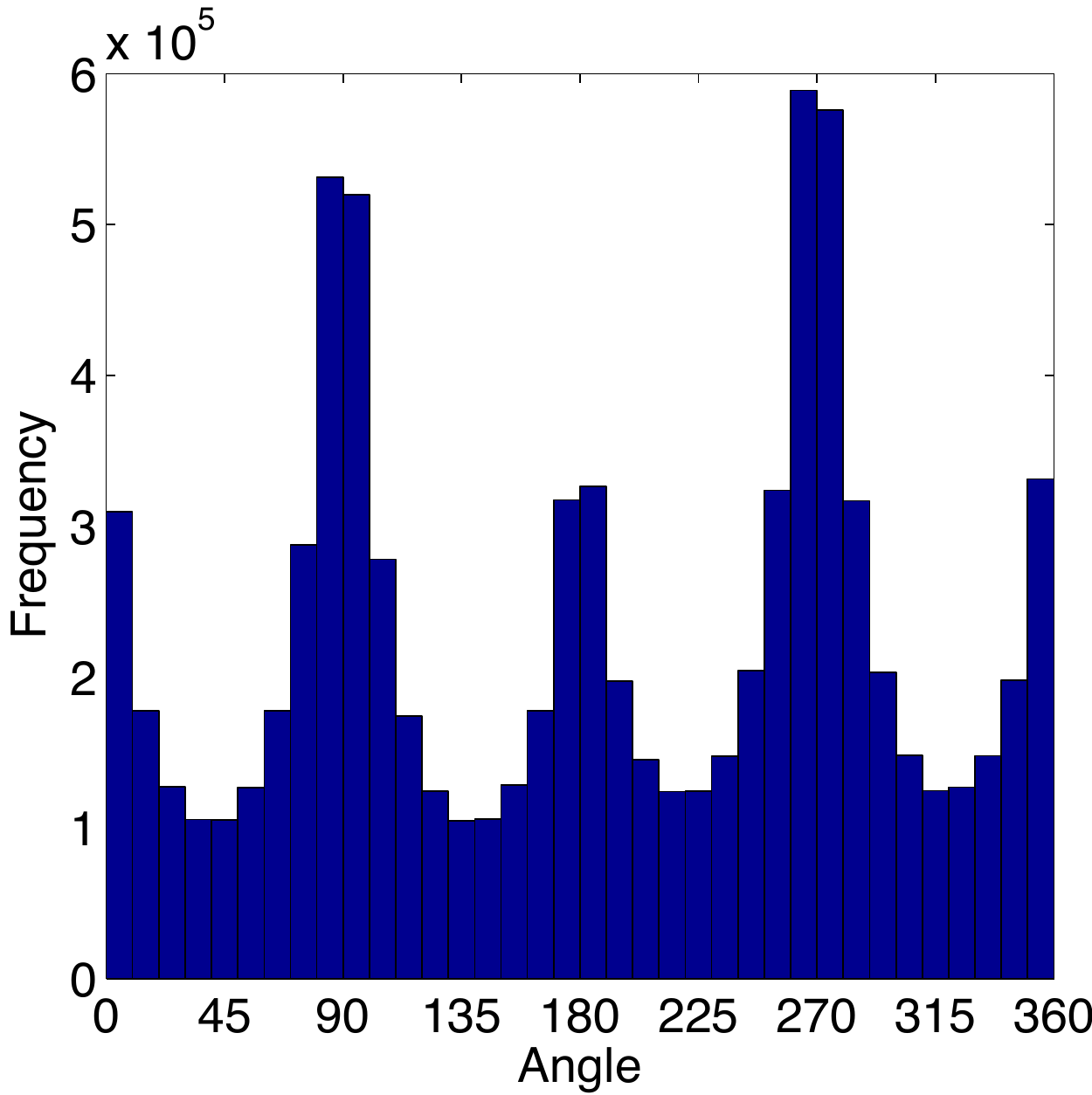}  &
\includegraphics[width=.47\linewidth]{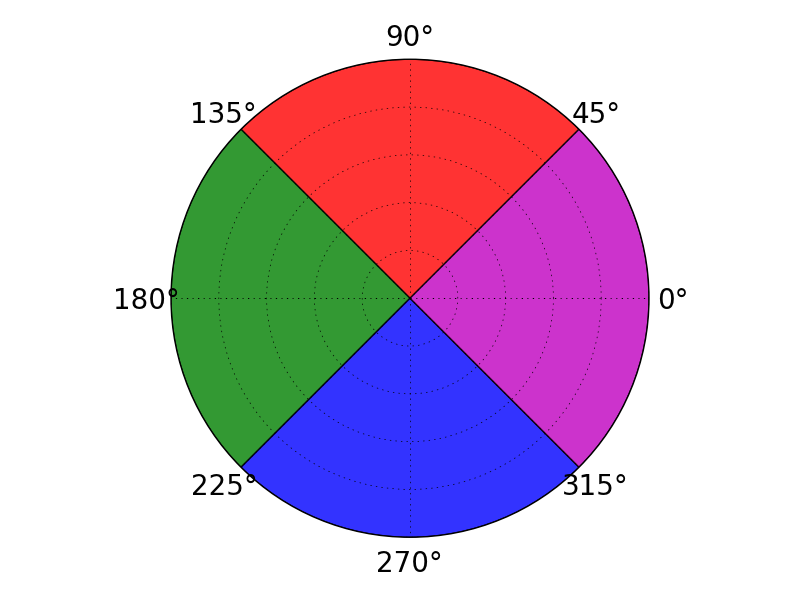} \\
(a) & (b)
\end{tabular}
\caption{(a) Distribution of keypoint angles from \textit{Holidays} dataset. (b) Learnt 4 angle bins with $\pi / 4$ offset.}
\label{fig:angleDistribution}
\vspace{-2ex}
\end{figure}

\begin{figure}[h]
\centering 
\epsfig{file=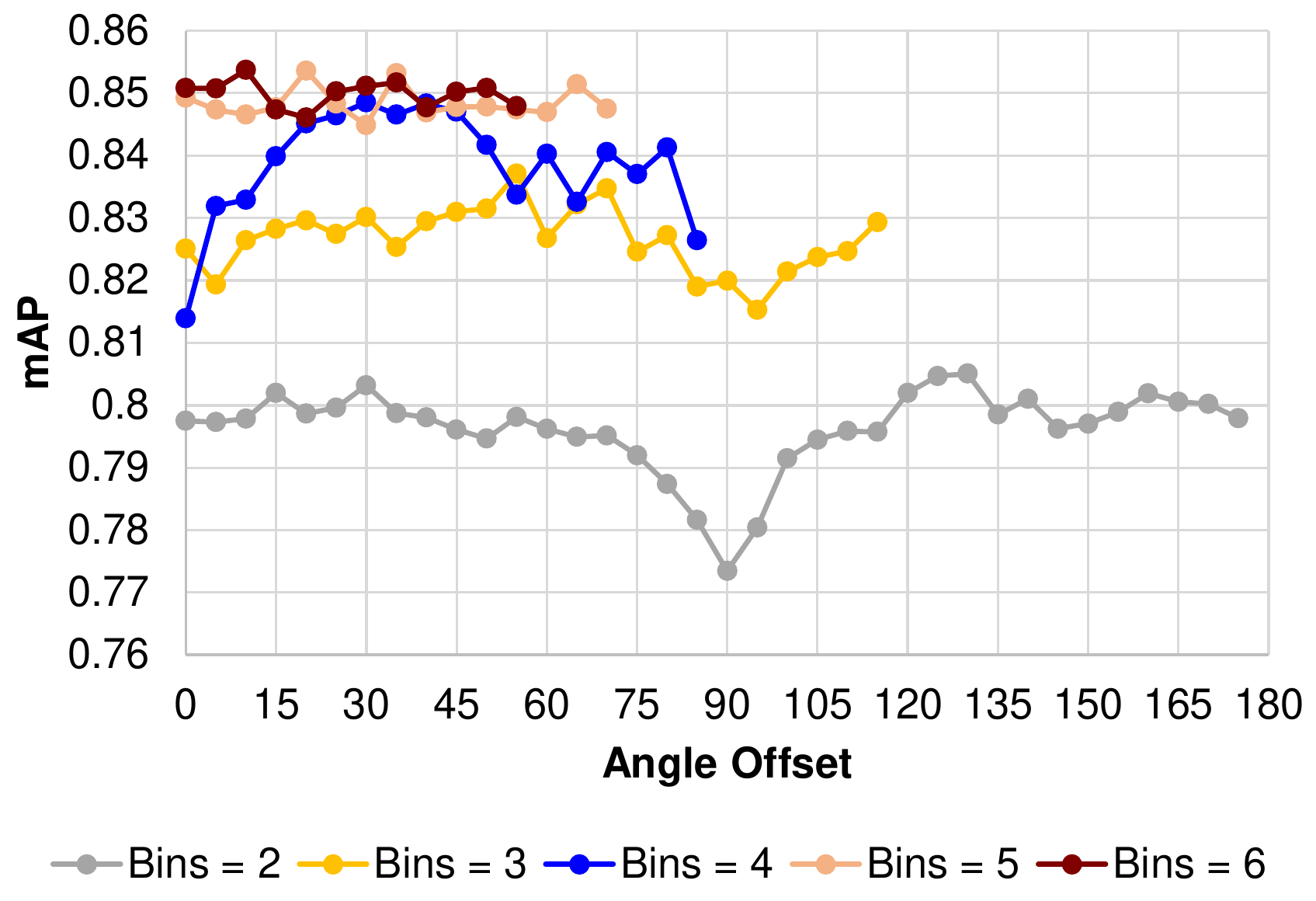,width=1.0\linewidth} 
\vspace{-0.3cm}
\caption{Performance using different angle bins and offset on \textit{Rotated Holidays} dataset. Based on empirical observation, we apply angle membership learning with $M=4$, and use the learned partition in our experiment. Such setting gives near best performance but lower dimension of gVLAD features as compared to using more bins for modeling angle distribution.} 
\label{fig:angle-compare}
\vspace{-1ex}
\end{figure}

\noindent\textbf{Angle Membership Function:}
The angle distribution of SURF keypoints from \textit{Holidays} dataset is shown in Figure~\ref{fig:angleDistribution} (a). We find that majority keypoints have vertical or horizontal angles as detectors have larger response at these points, resulting in roughly 4 centers ($\pi/2$, $\pi$, $2\pi/3$, $2\pi$).
To learn the membership function of each keypoint angle, we apply the proposed approach in ~\ref{sec:membership}. 
Because larger number of bins will increase the dimension of the final derived gVLAD feature, to gain a reasonable representation as well as low dimensionality, we set the number of angle bins to be $4$ to fit the distribution. A $\pi/4$ offset and a set of evenly distributed bins: $[-\pi/4, \pi/4)$, $[\pi/4, 3\pi/4)$, $[3\pi/4, 5\pi/4)$ and $[5\pi/4, 7\pi/4)$, are automatically estimated from the algorithm, which is visualized in Figure~\ref{fig:angleDistribution} (b). We use this angle bin partition in following evaluations.
We had also experimented using different number of bins and offset on \textit{Rotated Holidays}  dataset. We observe increasing performance as more bins are used, as shown in Figure~\ref{fig:angle-compare}. This is because that increasing number of bins is equivalent to increasing number of subspaces, by which the distance of descriptor towards centroid can be computed in a more discernible way. However, gains by using 5 or 6 bins as compared to the predicted angle partition ( 4 bins with $\pi/4$ offset) by propose algorithm are marginal, also our learned setting has much smaller dimensions.


\noindent\textbf{Vocabulary Generation:}
The vocabulary consisting of $K=256$ visual words is computed from all SURF descriptors on Paris dataset. Various different cluster initializations of $k$-means are executed and the best clustering is used as the vocabulary for all evaluations. As the number of extracted descriptors is typically much larger than $K$, e.g. even the smallest Holidays dataset contains 8.3 million SURF descriptors. This vocabulary can be considered independent from all datasets. Such simplification has been used in literature~\cite{arandjelovicall2013all}. For every dataset, this vocabulary is used as a reference vocabulary and a vocabulary adaptation is performed as described in section~\ref{sec:adaptation}.

\noindent\textbf{Retrieval:}
During retrieval, L2 distance is computed to rank images with respect to input query. Since our focus is generating a compact and efficient image descriptor, to illustrate the power of the proposed descriptor, we use brute-force distance computation to report our results. However, our proposed descriptors can in principle be used with approximate distance matching or other hashing based techniques as well, which is beyond the scope of this paper. 

\subsection{Performance Evaluation \& Analysis}
\label{sec:result}

The performance in all retrieval experiments is evaluated using the mean average precision (mAP), which is defined as the mean of the average precision over all the queries given a dataset. Average Precision is computed as the average of the precision value obtained for the set of top $k$ images after each relevant image is retrieved. We use standard evaluation packages obtained from the data websites.


\noindent\textbf{The Power of Angle:}
To illustrate the power of angle, we performed a simple experiment which uses only the angle information from each keypoint to retrieve similar images. After obtaining the keypoints and the angle of each keypoint, we generate an angle histogram for each of image by binning all angles into $Q$ bins. We use L2 distance to compute the similarity between angle descriptors. Table~\ref{tbl:only-angle-bin} shows the retrieval results on \textit{Rotated Holidays} dataset. It can be seen that surprisingly using only angle information (without any appearance information from SURF or other descriptors), we can still achieve about $26.9\%$ mAP results. Note that the dimension of the angle bin histogram for the best result is only $72$, which is a much smaller number compared to conventional BOW or VLAD descriptors. 
    
\begin{table}[t]\small
\centering
\begin{tabular}{ c | c | c | c | c | c | c }
				&\multicolumn{6}{c}{ \textbf{Angle Bins}}  \\ \hline
\textbf{$Q$} & 2 		  & 4        & 8        & 18      & 36       & 72	\\ \hline
\textbf{mAP} & 0.015 & 0.037 & 0.149 & 0.241 & 0.261 & 0.269 \\
\hline
\end{tabular}
\vspace{2mm}
\caption{Retrieval performance on \textit{Rotated Holidays} dataset using only Angle binning histogram with varying dimensions, no appearance information is used. Given only $72$ dimension of angle histogram, a surprising $26.9\%$ mAP result is achieved. }
\label{tbl:only-angle-bin}
\vspace{-2ex}
\end{table}

\begin{figure}[!ht]
\centering 
    \includegraphics[width=.45\textwidth]{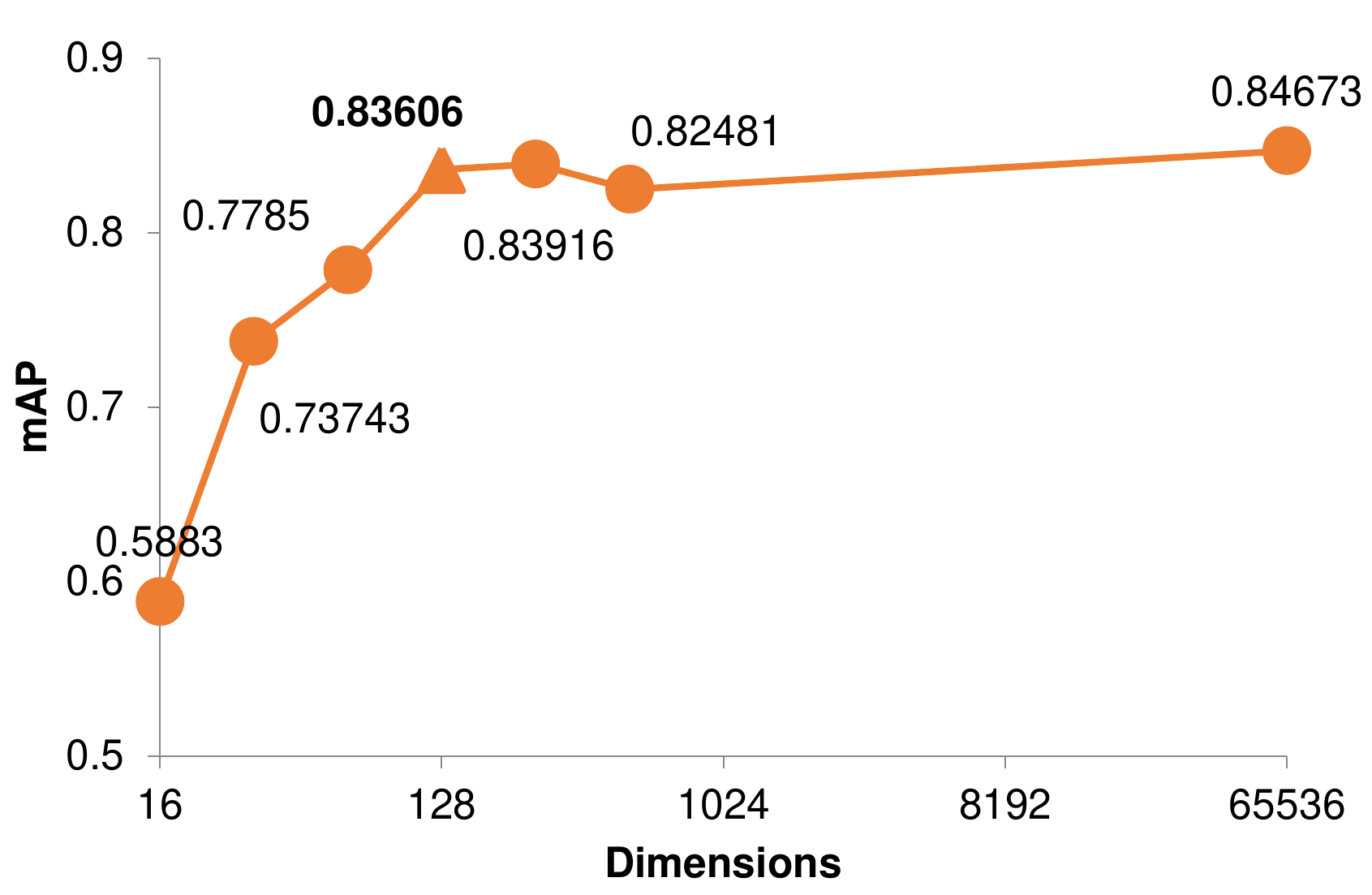}
  \vspace{-0.1cm}
  \caption{Dimension reduction on original gVLAD descriptor using PCA whitening. The original feature dimension is 65,536. After compressed to 128-D, the mAP decreases only about 1\%.}
  \label{fig:pca_compression}
\end{figure}

\noindent\textbf{Step-by-Step Performance Evaluation:}
To show the performance gain obtained from each of the proposed steps, we performed a step-by-step experiment on \textit{Rotated Holidays} dataset and baselined it with VLAD performance. Results are listed in Table~\ref{tbl:step-performance}. All results use SURF detector and SURF descriptor. It can be seen that by adding inter-Z-Score normalization to the original VLAD, the performance is increased by $5.4\%$. Performing Angle Binning over VLAD leads to a gain of $7.3\%$. By combining both Angle Binning and Z-Score normalization, we achieve $14.7\%$ improvement over VLAD representation. Performing vocabulary adaptation for \textit{Rotated Holidays} dataset provides additional $3.8\%$ performance gain. Finally, PCA whitening is applied which is able to reduce the dimension significantly with only about $1.1\%$ performance loss, as compared to PCA without whitening having a loss of $3.6\%$. To demonstrate the performance of low-dimensional gVLAD descriptor using PCA whitening, we further plot the mAP performance curve by varying $\rho=2^4$ to $2^{16}$ in Figure~\ref{fig:pca_compression}. It can be seen that with only $32$ dimensions, the performance by the proposed descriptor can reach to $mAP=0.737$, which already outperforms the original VLAD descriptor using $1024$ visual words with $65,536$ dimension ($mAP=0.670$) as shown in Table~\ref{tbl:step-performance}. 

We also test our proposed method using SIFT detectors and root SIFT~\cite{arandjelovic2012three} descriptors, since most previous published work use SIFT. For fair comparison, we implement VLAD with root SIFT descriptors, which have better performance compared with 0.526 on \textit{Holidays} dataset reported in~\cite{jegou2010aggregating}. Results as shown in Table~\ref{tbl:root_sift_vlad} demonstrate the superior performance of proposed approach over SIFT descriptors as well. As noted, comparing Tabel~\ref{tbl:root_sift_vlad}  and Table~\ref{tbl:step-performance}, we observe in general that using SURF descriptors outperforms SIFT based descriptors. 

\begin{table}[t]\small
\centering
\begin{tabular}{ c | c | c }
\textbf{Dataset} 	& \textbf{VLAD}* 		& \textbf{gVLAD} 	\\ \hline
Holidays 				& 0.548 									& \textbf{0.710}			\\ \hline
Rotated Holidays 	& 0.550 									& \textbf{0.786}\\  \hline
\end{tabular}
\vspace{1.5mm}
\caption{Comparison of our proposed gVLAD with VLAD on benchmark datasets. Both VLAD and gVLAD use SIFT detectors and root SIFT descriptors. $256$ visual words are used. The feature dimension of VLAD is $256\times 128 = 32768$, and for gVLAD is $256\times 128\times 4 = 131072$. * denotes our implementation.}
\label{tbl:root_sift_vlad}
\end{table}

\begin{table} \small
\centering
\begin{tabular}{c|c|c}
\textbf{Method} 													& \textbf{Dimension} 	& \textbf{mAP} \\ \hline
VLAD ($K=256$)* \cite{jegou2012aggregating}  	& 16,384 						& 0.662\\ \hline
VLAD ($K=1024$)*  \cite{jegou2012aggregating}  & 65,536 						& 0.670\\ \hline
VLAD ($K=256$) + inter-norm 							& 16,384 						& 0.716\\ \hline
VLAD ($K=256$) + Angle Binning 						& 65,536						 & 0.735\\ \hline
+ inter-norm 														& 65,536 						 & 0.809\\ \hline
+ Voc Adaptation 												& 65,536 						& \textbf{0.847}\\  \hline
PCA  																	& 128 							& 0.811\\ \hline
PCA + whitening													& 128 							& \textbf{0.836}\\ \hline
\end{tabular}
\vspace{2mm}
\caption{mAP on the \textit{\textbf{Rotated Holidays}} dataset comparing to start-of-art results. Best performances are in bold. *VLAD result in this table are based on our implementation. All results use SURF detector and SURF descriptor.}
\vspace{-2ex}
\label{tbl:step-performance}
\end{table}

\noindent\textbf{Full Size \& Low Dimensional gVLAD Descriptors:} 
We compared our proposed method with several benchmark results in~\cite{jegou2012aggregating,sivic2003video,perronnin2010large,arandjelovicall2013all} for both full size descriptor and dimension reduced descriptor ($\rho=128$). Experiments are done using both \textit{Holidays} dataset and \textit{Oxford 5K} dataset. Table~\ref{tbl:fullfeature} shows that the proposed approach significantly outperforms the state-of-the-art performance by approximately $16.6\%$ and $7\%$ on \textit{Holidays} and \textit{Oxford 5K} dataset respectively. For low dimensional case, as shown in Table~\ref{tbl:pcafeature}, our algorithm outperforms the best state-of-art result by $15\%$ on both datasets. 

Comparing Table~\ref{tbl:fullfeature}  and~\ref{tbl:pcafeature}, results also show that the proposed gVLAD descriptor is quite powerful in the sense that even with PCA whitening and reduced dimension, it can still achieve better result as compare to the best benchmark results with full size descriptors. In addition, PCA whitening based dimension reduction only results in small amount of performance decrease which is about $2.76\%$ in average of both datasets, and $1.1\%$ in the best case (\textit{Rotated Holidays}).  

\begin{table}\small
\centering
\begin{tabular}{c|c|c|c}
\textbf{Method}				& \textbf{Dimension} 					& \textbf{Holidays} 		& \textbf{Oxford} \\ \hline
BoW 20k-D \cite{jegou2012aggregating} \cite{sivic2003video} 		& 20,000 		& 0.452 & 0.354\\ \hline
BoW 200k-D \cite{jegou2012aggregating} \cite{sivic2003video} 	& 200,000 	& 0.540 & 0.364\\ \hline
Improved Fisher \cite{perronnin2010large} 									& 16,384 		& 0.626 & 0.418\\ \hline
VLAD \cite{jegou2010aggregating} 												& 8,192 		& 0.526 & -\\ \hline
VLAD + SSR \cite{jegou2012aggregating} 										& 16,384 		& 0.598 & 0.378\\ \hline
Improved VLAD + SSR  \cite{arandjelovicall2013all} 						& 32,768 		& - & 0.532\\ \hline
VLAD + intra-norm \cite{arandjelovicall2013all} 							& 32,768		& 0.646 & 0.555\\ \hline
Ours 																								& 65,536 		& \textbf{0.812} & \textbf{0.626}\\ \hline
\end{tabular}
\vspace{2mm}
\caption{mAP performance by full size gVLAD descriptors as compared to state-of-the-art results on \textit{\textbf{Holidays}} and \textit{\textbf{Oxford}}. Existing approaches are based on SIFT descriptors, while the proposed gVLAD descriptor uses SURF detector and SURF descriptor. Best performances are in bold.}
\vspace{-2ex}
\label{tbl:fullfeature}
\end{table}

\begin{table} \small
\centering
\begin{tabular}{c|c|c}
\textbf{Method} 						&  \textbf{Holidays} 				&  \textbf{Oxford}\\ \hline
GIST~\cite{jegou2012aggregating} 		 								& 0.365 & -\\ \hline
BoW ~\cite{jegou2012aggregating, sivic2003video}		 	& 0.452 & 0.194\\ \hline
Improved Fisher~\cite{perronnin2010large} 						& 0.565 & 0.301\\ \hline
VLAD~\cite{jegou2010aggregating} 									& 0.510 & -\\ \hline
VLAD + SSR~\cite{jegou2012aggregating} 							& 0.557 & 0.287\\ \hline
Multivoc-BoW~\cite{jegou2012negative} 								& 0.567 & 0.413\\ \hline
Multivoc-VLAD~\cite{jegou2012negative} 							& 0.614 & -\\ \hline
VLAD + intra-norm~\cite{arandjelovicall2013all} 				& 0.625 & 0.448\\\hline
Ours 																					& \textbf{0.779} & \textbf{0.600}\\ \hline
\end{tabular} \vspace{2mm}
\caption{mAP performance by gVLAD low dimensional descriptors ($\rho=128$): comparison with state-of-the-art on the \textit{Holidays} and \textit{Oxford 5k} benchmarks. The existing approaches are based on SIFT descriptors, while the proposed gVLAD descriptor uses SURF detector and descriptor. Best performances are in bold.}
\vspace{-2ex}
\label{tbl:pcafeature}
\end{table}

\noindent\textbf{Performance on Large Scale Dataset:} 
We scale the proposed algorithm to large scale image dataset with millions of images, and test on both using full size gVLAD and PCA dimension reduced 128-D descriptors. In total, $4$ large scale datasets are used, including \textit{Holidays + Flickr 1M}, \textit{Rotated Holidays + Flickr 1M}, \textit{Oxford 105K}, and \textit{Oxford 5K + Flickr 1M}. As can be seen from Table~\ref{tbl:large}, our methods outperform all current state-of-the-art methods. For example, using dimension reduced 128-D gVLAD descriptors, on \textit{Holidays + Flickr 1M} dataset, our algorithm outperforms the best result~\cite{arandjelovicall2013all} 	reported in literature with a significant gain of $22.8\%$. On \textit{Oxford 105K} dataset, we are able to achieve $11.6\%$ better result than ~\cite{arandjelovicall2013all}.

Further, same with our previous observation in Table~\ref{tbl:pcafeature} as compared to Table~\ref{tbl:fullfeature},  Table~\ref{tbl:large} also shows performance only drops very slightly using the proposed PCA whitening. This implies that the proposed gVLAD descriptor is quite powerful. Also, being combined with proper dimension reduction schema, effective representation with computational efficiency can be achieved.

\begin{table*} \small
\centering
\begin{tabular}{c|c|c|c|c|c}
							& \textbf{State of the Art} & \textbf{State of the Art} & \textbf{Ours} & \textbf{Ours} & \textbf{Ours}  \\
\textbf{Dataset} 	& \textbf{Original Dimension} & \textbf{128-D} & \textbf{Original Dimension} & \textbf{128-D} & \textbf{Loss in PCA} \\
\hline
\textit{Holidays} & 0.646 \cite{arandjelovicall2013all} & 0.625 \cite{arandjelovicall2013all} & \textbf{0.812} & \textbf{0.779} & 0.033\\
\hline
\textit{Holidays + Flickr 1M} & - & 0.378 \cite{arandjelovicall2013all} & - & \textbf{0.607} & -\\
\hline
\textit{Rotated Holidays} & - & - & \textbf{0.847} & \textbf{0.836}  & 0.011\\
\hline
\textit{Rotated Holidays + Flickr 1M} & - & - & - & \textbf{0.654} & -\\
\hline
\textit{Oxford 5K} & 0.555 \cite{arandjelovicall2013all} & 0.448 \cite{arandjelovicall2013all} & \textbf{0.626} & \textbf{0.600} & 0.026\\
\hline
\textit{Oxford 105K} & - & 0.374 \cite{arandjelovicall2013all} & - & \textbf{0.490}  & -\\
\hline
\textit{Oxford 5K + Flickr 1M} & - & - & - & \textbf{0.438} & -\\
\hline
\textit{Paris 6K} & 0.494 \cite{philbin2008lost} & - & \textbf{0.631} & \textbf{0.592} & 0.039 \\
\hline
\end{tabular} \vspace{2mm}
\caption{mAP Performance on large scale datasets: comparisons with benchmark results. Best performances are in bold.}
\vspace{-2ex}
\label{tbl:large}
\end{table*}

\subsection{Time Complexity and Memory Footprint}
\label{sec:complexity}

Each image takes 512 bytes in memory after being converted to 128 dimensional gVLAD feature vector by PCA compression. The largest dataset (\textit{Holidays + Flickr 1M}) in our experiment occupies $0.5$GB of RAM for keeping all features in memory. To evaluate the time complexity of each step in the proposed gVLAD computation, we conduct experiments on this dataset using a Ubuntu machine with two Xeon X5675 CPUs at 3.07GHz, with 12 physical cores and 24 logical cores in total. We rely on multi-threading whenever possible. Table~\ref{tbl:timing} illustrates the average results on $10$ randomly selected queries. As shown, our proposed technique takes approximately $100$ millisecond to compute gVLAD representation, and $750$ millisecond to perform an end-to-end brute-force retrieval over the entire inventory. Since our proposed descriptors can in principle be used with other approximate distance matching or indexing schema, better retrieval speed can be expected, which will be very useful in practical applications. 

\begin{table} \small
\centering
\begin{tabular}{c|c}
\textbf{Process} 								& \textbf{Mean $\pm$ std. (ms)} \\  \hline
SURF detection \& description 		& 373.5 $\pm$ 69.1 \\  \hline
gVLAD computation 						& 71.7 $\pm$ 20.3 \\ \hline
PCA compression 							& 28.0 $\pm$ 3.6 \\ \hline
Nearest neighbor search 					& 266.7 $\pm$ 36.3\\ \hline
\end{tabular}
\vspace{2mm}
\caption{Speed analysis based on 10 random query images from \textit{Holidays + Flickr 1M} dataset.}
\vspace{-2ex}
\label{tbl:timing}
\end{table}

\section{Conclusion}
\label{sec:conclusion}
We present gVLAD which is a novel extension of popular VLAD descriptor for large scale image search. Our proposed descriptor extends VLAD by integrating weak geometric cues in form of key point angles. A principled technique to represent this information as membership function over angles is also presented. The vocabulary adaptation and inter-Z-score normalization are also proposed to improve the performance of the system. Extensive experiments are conducted on existing publicly available benchmark datasets which demonstrate the superior performance of our approach. Our future work focuses on exploring efficient indexing strategies to avoid the brute-force matching of images. We are also investigating other related low level information that can be further integrated into gVLAD to make the representation more powerful.

{\small
\bibliographystyle{ieee}
\bibliography{gvladbib}
}

\end{document}